\documentclass[10pt,twocolumn,letterpaper]{article}

\usepackage{wacv}              %

\usepackage{amsmath}
\usepackage{amsfonts}
\usepackage{amssymb}
\usepackage{booktabs}
\usepackage{multirow}
\usepackage{gensymb}
\usepackage{adjustbox}

\definecolor{wacvblue}{rgb}{0.21,0.49,0.74}
\usepackage[pagebackref,breaklinks,colorlinks,allcolors=wacvblue]{hyperref}

\def\wacvPaperID{2952} %
\def\confName{WACV}
\def\confYear{2026}

\title{Adversarial Samples Are Not Created Equal}

\author{Jennifer Crawford\\
Scale AI\\
\and
Amol Khanna\\
CrowdStrike\\
\and
Fred Lu\\
Booz Allen Hamiltion\\
\and
Amy R. Wagoner\\
Booz Allen Hamiltion\\
\and
Stella Biderman\\
EleutherAI\\
\and
Andre T. Nguyen\\
Booz Allen Hamiltion\\
\and
Edward Raff\\
CrowdStrike\\
}

\begin{document}
\maketitle

\begin{abstract}
Over the past decade, numerous theories have been proposed to explain the widespread vulnerability of deep neural networks to adversarial evasion attacks. Among these, the theory of non-robust features proposed by Ilyas et al. \cite{bugs} has been widely accepted, showing that brittle but predictive features of the data distribution can be directly exploited by attackers. However, this theory overlooks adversarial samples that do not directly utilize these features. In this work, we advocate that these two kinds of samples - those which use use brittle but predictive features and those that do not - comprise two types of adversarial weaknesses and should be differentiated when evaluating adversarial robustness. For this purpose, we propose an ensemble-based metric to measure the manipulation of non-robust features by adversarial perturbations and use this metric to analyze the makeup of adversarial samples generated by attackers. This new perspective also allows us to re-examine multiple phenomena, including the impact of sharpness-aware minimization on adversarial robustness and the robustness gap observed between adversarially training and standard training on robust datasets.
\end{abstract}

\section{Introduction}

Deep neural networks have been widely noted to display a curious vulnerability to the manipulation of input data \cite{intriguing,raff_you_2025,doldo_stop_2025,raff_adversarial_2025,Richards2021,Rahnama2020,Raff_BaRT_2019,Nguyen2019_ANSR,Fleshman2018a}. In what has become known as adversarial evasion attacks, a malicious user can add engineered but imperceptible perturbations to otherwise normal inputs to trick a model into confidently outputting incorrect answers. These adversarial samples represent not only a glaring security risk but also betray the fact that neural networks have fundamental differences in their functioning compared to humans, in spite of their ability to match our performance on many tasks.

Following the discovery of adversarial samples, there have been many attempts to understand and characterize this behavior. For example, previous works have pointed to the high dimensionality of input data \cite{advinevitable}, overfitting \cite{tilt}, and the use of locally linear activation functions \cite{advlinear} as possible reasons for this vulnerability. In contrast, the seminal work of \cite{bugs} proposed that brittle but predictive features of the data which models use to solve tasks play a large role in adversarial susceptibility. That is, adversarial samples are a natural outcome of neural networks solving tasks differently from humans, by having a predisposition for learning the most simplistic input features for their task \cite{simplicitybias, starvation, shorcut}. Adversaries can directly manipulate these \textit{non-robust features} in order to change the meaning of the input from the model's perspective while leaving it unchanged for a human.

While this theory offers an intuitive explanation for the existence of adversarial samples, it does not directly address the possibility of adversarial samples that do not involve the explicit addition or removal of non-robust features. In fact, several works have brought into question whether brittle features of the data distribution offer a full explanation for adversarial vulnerability \cite{bugsrebut, dataalone}. In this work, we aim to differentiate these distinct weaknesses and examine their pervasiveness in adversarial attacks. Our contributions are as follows:
\begin{itemize}
    \item We propose an ensemble-based metric to help identify if an adversarial sample is directly utilizing non-robust features. We use this metric to distinguish \textit{adversarial bugs}, which we define as adversarial samples that do not rely on the addition or removal of non-robust features. 
    \item We examine the prevalence of non-robust features in adversarial attacks and find that the robustness of adversarially trained models breaks down when the perturbation strength is large enough to manipulate the predictive features of inputs. However, they continue to display a striking resilience to adversarial bugs. 
    \item We uncover a link between adversarial bugs and sharpness in the loss landscape and demonstrate that Sharpness-Aware Minimization (SAM) gives targeted protection against adversarial bugs. This is in contrast to recent work which suggest SAM's role in robustness stems from encouraging the learning of robust features.
    \item Using our metric, we re-examine robust datasets and find that, contrary to previous assumptions, they still contain non-robust features. This offers a clear explanation for the widespread robustness gap between adversarially trained models and models trained on robust datasets.
\end{itemize}

An outline of our paper is as follows. In Section \ref{sec:prelim}, we first provide relevant background into the theory of non-robust features. We then define and justify our metric to measure the manipulation of non-robust features in Section \ref{sec:method}. This metric is used to examine the composition of adversarial samples created by attacks on CIFAR10 and SVHN in standard and adversarially trained models (Section \ref{sec:comp}), models trained using SAM (Section \ref{sec:sam}), and models trained on robust datasets (Section \ref{sec:gap}).

\section{Preliminaries} \label{sec:prelim}

In this section, we describe our notation and define key terms. 

\paragraph{Non-Robust Features} In this paper, we focus on the multi-label classification task $\mathcal{X}\rightarrow\mathcal{Y}$ where $\mathcal{X}\in\mathbb{R}^{m\times{m}}$ and $\mathcal{Y}=\{1,..,k\}$. We adopt the definition for features proposed by \cite{bugs}, which defines these as functions of the form $\phi: \mathcal{X} \rightarrow \mathbb{R}^k$. Thus, features can be viewed as quantifying the presence of a particular task-relevant signal in an input sample $\boldsymbol{x}$.  In this paper, we further clarify that the features of relevance are features of the data, \textit{data features}, in that they are statistically correlated with the data distribution $\mathcal{D}$. Each feature can be characterized by its usefulness (i.e. predictive ability) and robustness against input perturbations. Following \cite{bugs}, we define a $\rho$-useful feature for classification as having the property:
\begin{align}
    \mathbb{E}_{(\boldsymbol{x},y) \sim \mathcal{D}}[\mathbb{I}_y \cdot \phi(\boldsymbol{x})] \ge \rho.
\end{align}
where $\mathbb{I}_y$ is a one-hot vector encoding the ground-truth label $y$. Additionally, a feature is $\gamma$-robustly useful if it maintains $\gamma$-usefulness under a set of valid perturbations, $\Delta$:
\begin{align}
    \mathbb{E}_{(\boldsymbol{x},y) \sim \mathcal{D}}[\text{inf}_{\boldsymbol{\delta} \in \Delta} (\mathbb{I}_y \cdot \phi(\boldsymbol{x}+\boldsymbol{\delta}))] \ge \gamma.
\end{align}

\cite{bugs} demonstrated the impact of non-robust and robust features on adversarial susceptibility by creating non-robust and robust datasets, denoted here as $D_\mathit{NR}$ and $D_\mathit{R}$, which both result in generalization to the original data distribution $\mathcal{D}$. To create $D_\mathit{NR}$ and $D_\mathit{R}$, \cite{bugs} paired each input in the original train set with a separate ``target'' input (also in the original train set) and performed gradient descent on a standard or adversarially trained model, respectively, in order to optimize  
\begin{align}
    \text{min}_{\boldsymbol{x}'}\|g(\boldsymbol{x}_{target}) - g(\boldsymbol{x}')\|_2
    \label{eqn:dataset_gen}
\end{align}
where $\boldsymbol{x}_\mathit{target}$ is the target input and $\boldsymbol{x}'$ is initialized as $\boldsymbol{x}'_0 =\boldsymbol{x}$. Here, $g$ denotes the model's penultimate hidden representation. The output of this optimization is then paired with the label of the target input, $y_\mathit{target}$, to form a new sample in $D_\mathit{NR}$ or $D_\mathit{R}$. The authors showed that standard training only achieved nontrivial robustness against adversarial examples when using $D_R$.

For completeness, we reproduce this experiment in Appendix \ref{app:replication}. In agreement with \cite{bugs}, we find that standard training on $D_\mathit{NR}$ results in good test performance in spite of its inputs appearing mislabeled. Moreover, the robust accuracy is comparable to that of the original model. In contrast, we find that standard training on $D_\mathit{R}$ yields both nontrivial test and robust accuracy. However, similar to \cite{bugsrebut, dataalone}, we note a drop in robust accuracy (61.6\% $\rightarrow$ 2.0\% for $\epsilon=8/255$) as compared to adversarial training. We probe this robustness gap in Section \ref{sec:gap}.

In addition, \cite{bugs} formed two non-robust datasets, $D_\mathit{rand}$ and $D_\mathit{det}$, consisting of adversarial samples created via Projected Gradient Descent (PGD) \cite{pgd} using either a random or deterministic target label, $y_\mathit{target}$, respectively. To create $D_\mathit{det}$, we follow \cite{bugs} and set $y_\mathit{target}=(y+1)$ mod $C$, where $C$ denotes the number of classes. Each adversarial sample is paired with its target label, such that only non-robust features are predictive of the input label in $D_\mathit{rand}$, whereas both non-robust and robust features are available for learning in $D_\mathit{det}$.\footnote{Feature robustness here is in reference to the strength of perturbation applied during the PGD attack that generates $D_\mathit{rand}$ and $D_\mathit{det}$.} However, the robust features in $D_\mathit{det}$ are in direct opposition to solving the original test task. As replicated in Appendix \ref{app:replication}, training on both $D_\mathit{rand}$ and $D_\mathit{det}$ yields nontrivial generalization to the test distribution. This reveals that non-robust features are capable of being learned, even when paired with misaligned robust features as in $D_\mathit{det}$.

\begin{figure}[h!]
\centering
\includegraphics[width=0.6\columnwidth]{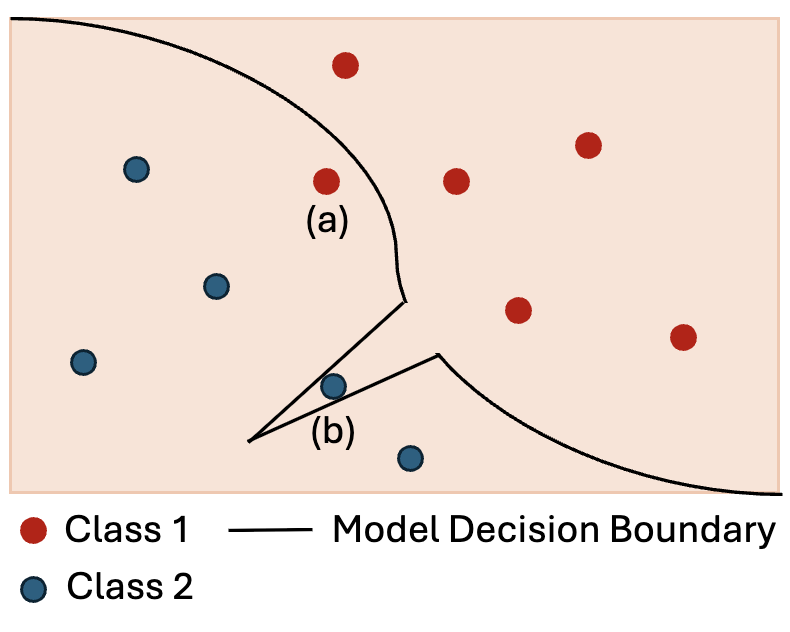}
\caption{\small{Illustration of the decision boundary for a deep neural network, trained on the task of binary classification, that allows for (a) adversarial samples that utilize non-robust features and (b) adversarial bugs. The former exist because the classifier used predictive features of the data distribution that are not human-aligned, while the latter appear as irregular blind spots of a particular model instance.}}
\label{fig:mental_img}
\end{figure}

\paragraph{Adversarial Bugs} The theory put forth by \cite{bugs} of non-robust features does not address the possibility of adversarial samples that do not directly utilize these features \cite{advinevitable, advlinear, tilt, discuss}. In this paper, we term such samples \textit{adversarial bugs} as they are not directly caused by adversarial perturbations that add or remove data features. Rather, they appear to be bugs or blind spots of a particular instance of model weights. An illustration of the distinction between adversarial bugs and adversarial samples that are crafted using non-robust features is shown in Figure \ref{fig:mental_img}. In this paper, we aim to empirically examine these two flavors of adversarial samples and their prevalence during adversarial attacks.

\section{Distinguishing Adversarial Bugs} \label{sec:method}

\subsection{Properties of Data Features}

In order to discern adversarial samples that utilize statistical data features, we first describe two desirable properties that we expect from an idealized model for such inputs.

\paragraph{Generalization Property} Training on adversarial samples that strongly exploit data features, when paired with their adversarial target label, should lead to non-trivial performance on the original test set. This was first observed in \cite{bugs}. We note that perfect generalization is not guaranteed, due to distributional shift as well as the fact that adversarial perturbations may not necessarily use a sufficient diversity of features.

\paragraph{Invariance Property} When applying random data augmentation during training, the model is highly encouraged to use features that are invariant to the chosen augmentations, $T$ (i.e. $f(\boldsymbol{x}) = f(T(\boldsymbol{x}))$). Thus, we expect adversarial samples that utilize data features to be largely invariant to these transformations (i.e. invariance comparable to that of natural test samples).

\subsection{Metric for Data Features}

We desire a metric that can measure the usage of data features in the creation of an adversarial sample, $\boldsymbol{x}_\mathit{adv}$, from its source input, $\boldsymbol{x}_\mathit{src}$. For this purpose, we note that for many neural network architectures and tasks it has been observed that different random initializations of a model will converge to very similar decision boundaries \cite{learntwice, seedvar, pythias}. This conveys the use of the same predictive input features and suggests that the transferability of adversarial samples among model instances may correlate with the manipulation of data features by adversarial perturbations. Therefore, we propose to analyze adversarial samples from the perspective of an ensemble of model instances initialized from different seeds.

\paragraph{Assumption 3.2} Let $f_{\theta_0}: \mathcal{X}\rightarrow \mathbb{R}^k $ denote a model instance, where $\theta_0$ are the neural network's parameters which follow the distribution $p(\theta)$. To approximate $\mathbb{E}_{\theta}[f_{\theta}]$, we use an ensemble of $N$ model instances $\tilde{f}(\boldsymbol{x}) = \frac{1}{N}\sum_{i=0}^{N-1}f_i(\boldsymbol{x})$ where $f_i$ denotes the $i$-th member of the ensemble. For a given source-adversarial input pair $(\boldsymbol{x}_\mathit{src}, \boldsymbol{x}_\mathit{adv})$ of $f_{\theta_0}$,  significant divergence of $f_\mathit{ens}(\boldsymbol{x}_\mathit{src})$ and $f_\mathit{ens}(\boldsymbol{x}_\mathit{adv})$ indicates that the adversarial perturbation has significantly manipulated data features. Conversely, agreement between these outputs conveys that the attacker largely relied on other adversarial weaknesses in the creation of $\boldsymbol{x}_\mathit{adv}$.

For each source-adversarial input pair of $f_{\theta_0}$, we propose calculating the Jensen–Shannon distance between the ensemble outputs, $JS(\tilde{f}(\boldsymbol{x}_\mathit{adv}) \| \tilde{f}(\boldsymbol{x}_\mathit{src}))$. For adversarial samples that alter data features, we expect this quantity to be large as it signifies that the meaning of the input has changed from the perspective of $\tilde{f}$. However, this quantity doesn't take into account a model's scale for the Jensen-Shannon distance between inputs of different labels, as a model's calibration will impact this distance.\footnote{For example, a model that is more confident in its predictions will naturally give larger Jensen-Shannon distances between the model outputs of two inputs with different labels.}  Therefore, we propose using the following normalized metric:
\begin{align}
    JS_{\Delta}(\boldsymbol{x}_\mathit{src}, y_\mathit{src}, \boldsymbol{x}_\mathit{adv}, y_\mathit{adv}) \equiv \frac{JS(\tilde{f}(\boldsymbol{x}_\mathit{src})\| \tilde{f}(\boldsymbol{x}_\mathit{adv}))}{\rho_{y_\mathit{src}, y_\mathit{adv}}}
    \label{eqn:kl_metric}
\end{align}
where $\rho_{y_\mathit{src}, y_\mathit{adv}} = {\mathbb{E}[JS(\tilde{f}(\boldsymbol{x}_{1})\| \tilde{f}(\boldsymbol{x}_2))]}$ for $\{(\boldsymbol{x}_1, y_1), (\boldsymbol{x}_2, y_2) \sim \mathcal{D} | (y_1, y_2)=(y_\mathit{src}, y_\mathit{adv})\}$. For example, a value of $JS_\Delta = 1$ would indicate that $\tilde{f}$ views the adversarial sample and its source input as meaningfully different -- comparable to a pair of clean inputs with labels $(y_\mathit{src}, y_\mathit{adv})$. Inversely, a value of $JS_\Delta \sim 0$ indicates that the adversary has left the data features of the input largely unchanged such that the adversarial sample is a bug particular to the attacked model instance. Thus, for any source-adversarial input pair, we use Equation \ref{eqn:kl_metric} to inspect the usage of data features in the creation of $\boldsymbol{x}_\mathit{adv}$. 

We now justify $JS_\Delta$ by confirming that transferability of adversarial samples to $\tilde{f}$ is correlated with the Generalization and Invariance properties of data features.

\paragraph{$JS_\Delta$ and the Generalization Property} In order to link $JS_\Delta$ and the Generalization property, we note that \cite{discuss} performs an experiment in which an attacker explicitly creates adversarial samples that do not transfer to a separate ensemble of identically trained models.  When using these highly non-transferable samples to form a dataset akin to $D_\mathit{det}$, they observe that training on this new dataset does not result in generalization to the original test distribution, $D$. We reproduce this experiment in Appendix \ref{app:gen_exp} and confirm that the transferability of adversarial samples to $\tilde{f}$ is correlated with the Generalization property.

\paragraph{$JS_\Delta$ and the Invariance Property} Lastly, we examine the relationship of $JS_\Delta$ to the Invariance property. For this experiment, we create a set of adversarial test samples, $X_\mathit{adv}$, using a targeted PGD-100 attack of $\ell_\infty$-constrained perturbations with $\epsilon=16/255$. We randomly select target labels, $Y_\mathit{target}$, such that they do not match the labels of the source inputs, $Y_{src}$. In addition, we also create a set of adversarial test samples that are highly non-transferable, $X^{\perp}_{adv}$, using the method proposed by \cite{discuss} (described here in Appendix \ref{app:gen_exp}). In Table \ref{tab:t_invariance}, we inspect the invariance of $f_{\theta_0}(\boldsymbol{x})$ to the random data augmentations, $T$, that were used to train the network and contrast $\boldsymbol{x} \sim X_\mathit{adv}$ with $\boldsymbol{x} \sim X^{\perp}_\mathit{adv}$. In this experiment, we utilize a standard augmentation strategy during training that includes random horizontal flips, random cropping, small random rotations, and a random color jitter. Note that these random augmentations are not applied as a defense during generation of $X_\mathit{adv}$ and $X^{\perp}_\mathit{adv}$. We advance the work of \cite{discuss} by determining that samples generated by vanilla PGD are largely invariant to $T$, while those that are highly non-transferable to $\tilde{f}$ (therefore having small $JS_\Delta$) are overwhelmingly not. In fact, when applying the train augmentations to these samples, we find that they are often no longer adversarial, with the attacked model correctly outputting their ground-truth label in the majority ($\sim86\%$) of cases.\footnote{This result suggests an alternative metric to help distinguish adversarial bugs from those that use non-robust features: invariance to the random data transformations applied during training. However, this not only requires the use of augmentations during training but also is highly dependent upon its strength.} This indicates that a large portion of the adversarial success of these samples stem from mechanisms distinct from the direct manipulation of data features.

\begin{table}[!h]
\caption{\small{Invariance of adversarial test samples to the random data augmentations $T$ used during training. We note that the majority of samples generated by vanilla PGD, $X_\mathit{adv}$, display high invariance, maintaining their target label $\sim95\%$ of the time under $T$. The opposite is true for samples that are highly non-transferable, $X^{\perp}_\mathit{adv}$, in which $\sim86\%$ of samples revert to the label of their source image under $T$.}}
\label{tab:t_invariance}
\centering
\begin{tabular}{ccc}
\toprule
\multirow{2}{*}{Input} & \multicolumn{2}{c}{Accuracy} \\
 & $Y = Y_\mathit{target}$ & $Y = Y_\mathit{src}$ \\
\midrule
$X_\mathit{adv}$ &  100.0\% & 0.0\% \\
$T(X_\mathit{adv})$ & 95.1\% & 4.5\% \\
$X^{\perp}_\mathit{adv}$ & 98.1\% & 2.4\%\\
$T(X^{\perp}_\mathit{adv})$ &  3.9\% & 85.7\%\\
\bottomrule
\end{tabular}
\end{table}

\paragraph{Remark} We note that adversarial bugs and those that use data features likely exist on a continuous spectrum. Even when using Equation \ref{eqn:ens_pgd} to directly target adversarial bugs, non-robust feature leakage may still occur in the creation of these samples and could, in practice, be difficult to fully remove. In addition, there are likely architectures and datasets for which Assumption 3.2 will not hold, particularly in cases where there is significant performance variability across training runs. %

\begin{figure*}[!h]
\centering
\begin{tabular}{ccc}
    \small $\epsilon = 3/255$ &
    \small $\epsilon = 5/255$ &
    \small $\epsilon = 8/255$ \\
    \includegraphics[width=0.3\textwidth]{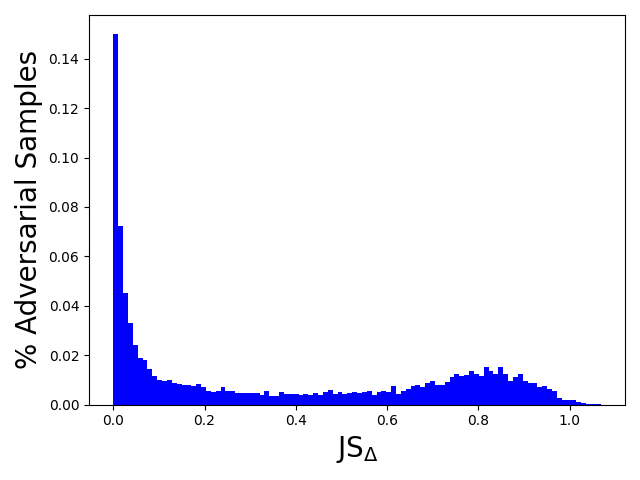} &
    \includegraphics[width=0.3\textwidth]{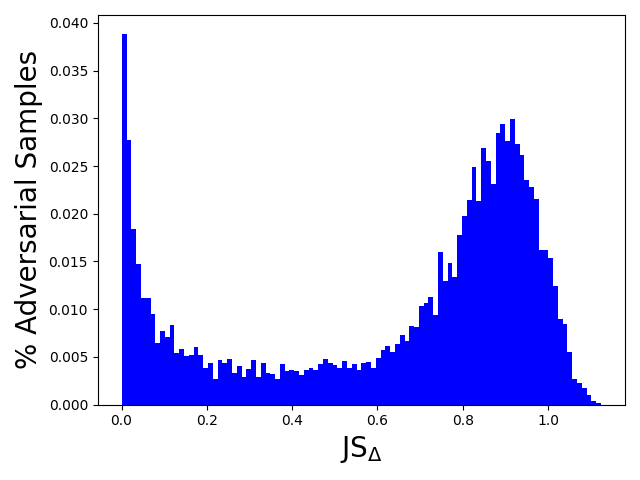} &
    \includegraphics[width=0.3\textwidth]{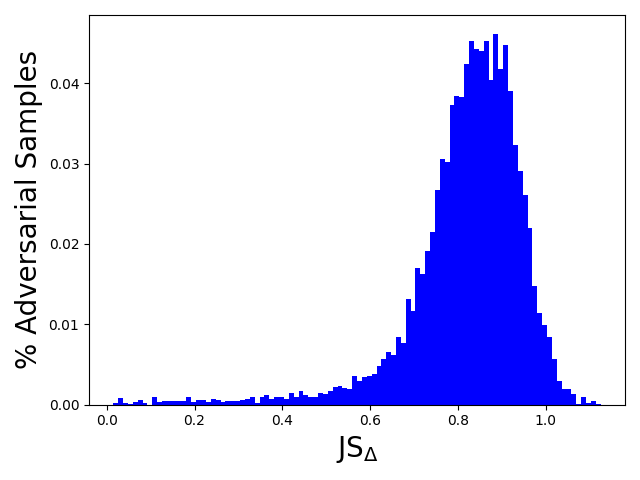} \\
\end{tabular}
\caption{\small{Normalized histograms of $JS_\Delta$ for adversarial samples of a non-robust ResNet50 model trained on CIFAR10. All samples are generated via a targeted PGD-100 attack. We observe that for low magnitudes of perturbation, a large percentage of adversarial samples can be identified as adversarial bugs. For larger magnitudes of perturbation, the attacker is able to create manipulate non-robust features in the majority of adversarial samples.}}
\label{fig:cifar_hist}
\end{figure*}

\begin{figure*}[!h]
\centering
\begin{tabular}{ccc}
    \small $\epsilon = 8/255$ &
    \small $\epsilon = 16/255$ &
    \small $\epsilon = 32/255$ \\
    \includegraphics[width=0.3\textwidth]{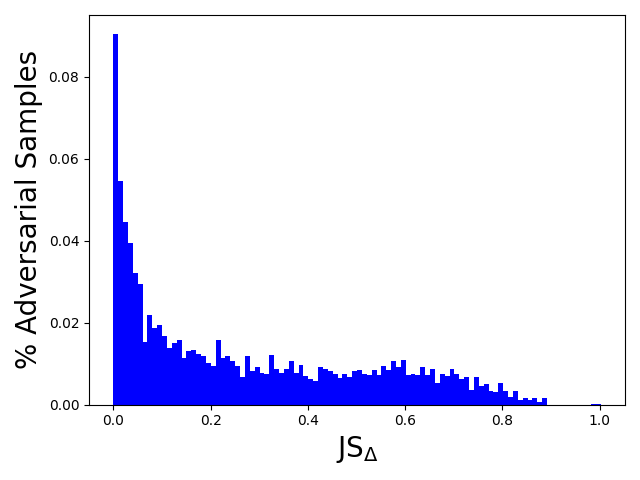} &
    \includegraphics[width=0.3\textwidth]{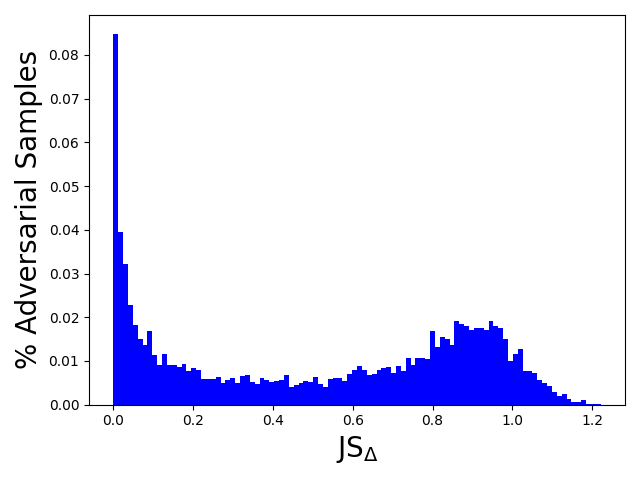} &
    \includegraphics[width=0.3\textwidth]{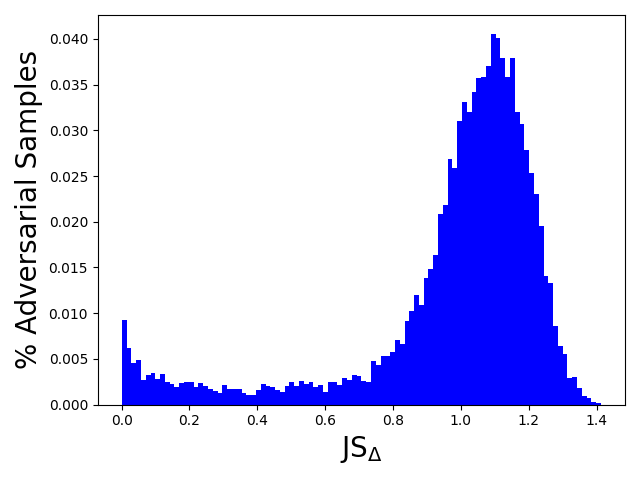} \\
\end{tabular}
\caption{\small{Normalized histograms of $JS_\Delta$ for successful adversarial samples of a robust ResNet50 model adversarially trained on CIFAR10. All samples are generated via a targeted PGD-100 attack. We observe that larger attack perturbations are needed to create both adversarial bugs and adversarial samples that utilize non-robust features.}}
\label{fig:cifar_adv_hist}
\end{figure*}

\section{Composition of Adversarial Samples} \label{sec:comp}

\paragraph{Experimental Setup} Similar to \cite{bugs}, we use ResNet50 and ResNet18 architectures \cite{resnet} trained on CIFAR10 \cite{cifar} and SVHN \cite{svhn}, respectively.\footnote{Convolutional architectures were found in \cite{learntwice} to have high decision boundary similarity among random seeds, which supports the application of Assumption 3.2.} Training details for our models are provided in Appendix \ref{app:training}. To estimate $\tilde{f}$ for each attacked model, $f_0$, we train a set of four identical model instances initialized from different random seeds. We note that in each experiment, the members of $\tilde{f}$ are trained using the same optimization procedure as for $f_0$. For our attack, we utilize PGD \cite{pgd} with 100 steps of $\ell_{\infty}$-constrained perturbations using a targeted cross-entropy attack loss and step size given by $\epsilon/10$. Due to space constraints, we provide the results for SVHN in Appendix \ref{app:svhn}. Note that in all of our experiments where we examine the composition of adversarial samples, we only consider \textit{successful} adversarial samples that result in a label change for the attacked model.%

First, we inspect the composition of adversarial samples for non-robust models trained using SGD with cross-entropy. In Figures \ref{fig:cifar_hist} and \ref{fig:svhn_hist}, we record a histogram of $JS_{\Delta}$ for each adversarial test sample of CIFAR10 and SVHN. We provide a quantitative analysis of these histograms in Tables \ref{tab:cifar_comp} and \ref{tab:svhn_comp}, respectively, where we calculate the percentage of adversarial samples that have $JS_\Delta < \beta$ for $\beta=[0.01, 0.05, 0.10]$. We find that for small magnitudes of attack a sizable percentage of adversarial samples are adversarial bugs and are highly non-transferable to $\tilde{f}$. After a certain threshold of attack perturbation is reached, the attacker switches to mainly utilizing non-robust features. This indicates that 1) the magnitude of perturbation needed to create adversarial bugs is often smaller than what is needed to manipulate non-robust features and 2) gradient-based attacks favor adversarial samples that manipulate data features. This latter point is intuitive as we expect the predictive features of the data to be a primary signal for shaping the loss landscape throughout training and so these have a strong influence during gradient-based attacks.

We repeat these experiments with adversarially trained models in Figures \ref{fig:cifar_adv_hist} and \ref{fig:svhn_adv_hist} and Tables \ref{tab:cifar_adv_comp} and \ref{tab:svhn_adv_comp} for CIFAR10 and SVHN. For both datasets, we observe a similar pattern of non-robust features being the favored mode of adversarial attack past a certain threshold of perturbation magnitude. We find that a notably higher magnitude of perturbation is needed to sufficiently manipulate data features (i.e. $32/255$ and $16/255$ for CIFAR10 and SVHN, respectively), as well as note an overall protection against adversarial bugs at small perturbation magnitudes. Thus, it appears that while adversarially training encourages models to learn more robust features it also provides protection against adversarial bugs.

\begin{table}[!h]
\caption{\small{Percentage of adversarial samples with $JS_\Delta<\beta$ for $\beta=[0.01, 0.05, 0.10]$ for a non-robust ResNet50 model trained on CIFAR10. All samples are generated via a targeted PGD-100 attack. We find that weak attacks tend to utilize adversarial bugs whereas a significant portion of adversarial samples use non-robust features beyond a certain threshold of attack perturbation.}}
\label{tab:cifar_comp}
\centering
\begin{tabular}{ccccc}
\toprule
\multirow{2}{*}{\begin{tabular}[c]{@{}c@{}}Attack \\ strength ($\epsilon$)\end{tabular}} & \multicolumn{1}{c}{\multirow{2}{*}{\begin{tabular}[c]{@{}c@{}}Robust \\ Accuracy\end{tabular}}} & \multicolumn{3}{c}{\% samples with $JS_\Delta<$}                                  \\ \cmidrule(l){3-5} 
                                                                                         & \multicolumn{1}{c}{}                                                                            & \multicolumn{1}{c}{$0.01$} & \multicolumn{1}{c}{$0.05$} & \multicolumn{1}{c}{$0.10$} \\
\midrule
1 / 255                                                                                  & 66.5\%                                                                                          & 23.1\%                     & 51.4\%                       & 64.3\%                     \\
3 / 255                                                                                  & 3.7\%                                                                                           & 14.3\%                     & 32.3\%                       & 39.6\%                     \\
5 / 255                                                                                  & 0.7\%                                                                                           & 0.3\%                      & 10.4\%                       & 14.5\%                     \\
8 / 255                                                                                  & 0.7\%                                                                                           & 0.0\%                      & 0.1\%                        & 0.2\%                      \\ \bottomrule
\end{tabular}
\end{table}

\begin{table}[!h]
\caption{\small{Percentage of adversarial samples with $JS_\Delta<\beta$ for $\beta=[0.01, 0.05, 0.10]$ for a robust ResNet50 model adversarially trained on CIFAR10. All samples are generated via a targeted PGD-100 attack. We observe that larger attack perturbations are needed to create both adversarial bugs and adversarial samples that utilize non-robust features.}}
\label{tab:cifar_adv_comp}
\centering
\begin{tabular}{ccccc}
\toprule
\multirow{2}{*}{\begin{tabular}[c]{@{}c@{}}Attack \\ strength ($\epsilon$)\end{tabular}} & \multicolumn{1}{c}{\multirow{2}{*}{\begin{tabular}[c]{@{}c@{}}Robust \\ Accuracy\end{tabular}}} & \multicolumn{3}{c}{\% samples with $JS_\Delta<$}                                  \\ \cmidrule(l){3-5} 
                                                                                         & \multicolumn{1}{c}{}                                                                            & \multicolumn{1}{c}{$0.01$} & \multicolumn{1}{c}{$0.05$} & \multicolumn{1}{c}{$0.10$} \\
\midrule
5 / 255 & 78.8\% & 9.0\% & 29.8\% & 42.8\% \\
8 / 255 & 61.6\% & 9.0\% & 26.1\%& 36.6\%\\
16 / 255 & 14.2\% & 7.5\% & 18.1\%& 24.5\%\\
32 / 255 & 1.4\% & 0.7\% & 2.3\%& 3.5\%\\
\bottomrule
\end{tabular}
\end{table}

\section{Connection to Sharpness-Aware Minimization (SAM)} \label{sec:sam}

\begin{figure*}[!h]
\centering
\begin{tabular}{ccc}
    \includegraphics[width=0.3\textwidth]{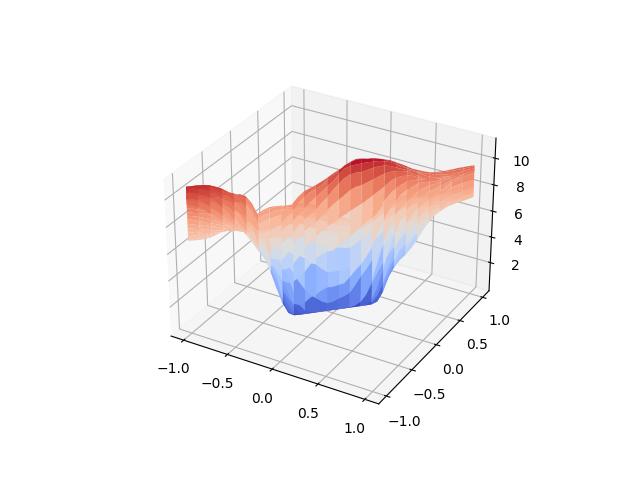} &
    \includegraphics[width=0.3\linewidth]{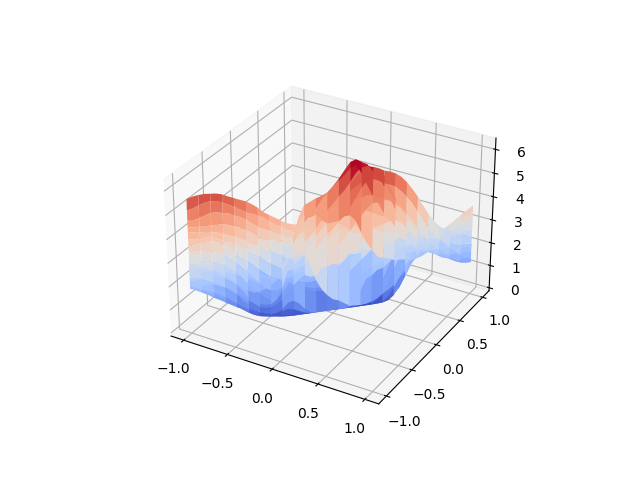} &
    \includegraphics[width=0.3\linewidth]{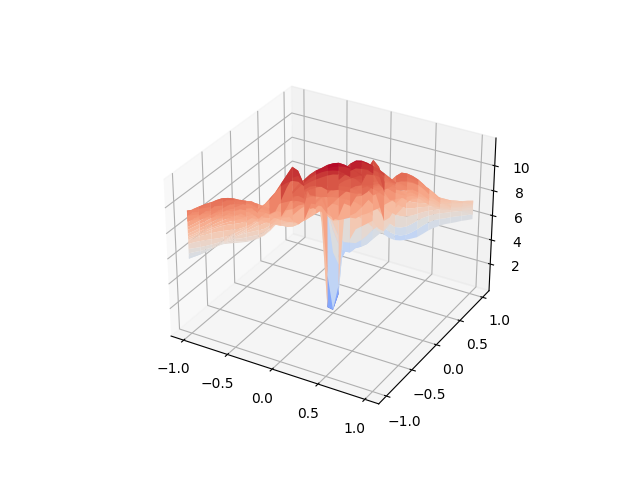} \\
  \end{tabular}
\caption{\small{Loss landscape surrounding a natural test sample (left) and adversarial test samples with large (middle) and small (right) $JS_\Delta$ for a ResNet50 model trained on CIFAR10. We find that the loss of adversarial bugs (right) displays a sharp minimum, while the loss for adversarial samples that manipulate data features (middle) have curvature comparable to that of natural samples.}}
\label{fig:surface}
\end{figure*}

In Section \ref{sec:method} we confirmed that the transferability of adversarial samples is correlated with their invariance to random data augmentations. Hence, it is natural to suspect that adversarial bugs have notably lower stability to random perturbations than clean inputs or adversarial samples that utilize data features. This property of adversarial bugs would correspond to residing in areas of low curvature, a trait that was separately linked to adversarial transferability \cite{boostreverse, boostflat}. To confirm our hypothesis, we inspect the loss landscape surrounding adversarial samples with high/low $JS_\Delta$ and natural test samples. Specifically, we plot $L_{CE}(\boldsymbol{x} + \alpha_1\boldsymbol{\epsilon_1} + \alpha_2\boldsymbol{\epsilon_2}, y)$ with $\alpha_1$, $\alpha_2$ $\in$ [-1,1] for two random directions $\boldsymbol{\epsilon_1}$, $\boldsymbol{\epsilon_2}$. For natural test samples, we explore the loss wrt. their ground-truth label, while we use the adversarial target label for samples altered by the attacker. In Figure \ref{fig:surface}, we show example surface plots of the loss landscape and find that the minima of adversarial bugs are notably sharper than minima that correlate with data features.

The insight that adversarial bugs exist in areas of high curvature might lead to the speculation that their sharpness is directly related to their mechanism. Recently, \cite{duality} made a comparison between Sharpness-Aware Minimization (SAM) and adversarial training. SAM can be formulated as the following min-max optimization problem:
\begin{align}
\text{min}_{\boldsymbol{w}} \mathbb{E}_{(\boldsymbol{x}, y) \sim D}[\text{max}_{||\boldsymbol{\epsilon}|| < \rho}L(\boldsymbol{w} + \boldsymbol{\epsilon}, \boldsymbol{x}, y)].
\end{align}
where perturbations are added to the model weights, $\boldsymbol{w}$, instead of the input, $\boldsymbol{x}$, as performed in adversarial training. Here, $\rho$ is a hyperparameter that limits the magnitude of perturbation. The authors of \cite{duality} took note of the similarity between the formulations of SAM and adversarial training and speculated that minimizing curvature might encourage a model to learn more robust features. In the study, models trained with SAM were observed to have adversarial robustness for small attack perturbations, without an accompanying drop in clean accuracy. This was used as empirical support for the hypothesis that SAM plays a role similar to adversarial training and inhibits non-robust feature learning. Using our transferability metric for adversarial samples, we investigate this claim.

\begin{figure*}[!h]
\centering

  \begin{tabular}{ccc}
  \small $\epsilon = 3/255$ &
    \small $\epsilon = 5/255$ &
    \small $\epsilon = 8/255$ \\
    \includegraphics[width=0.30\textwidth]{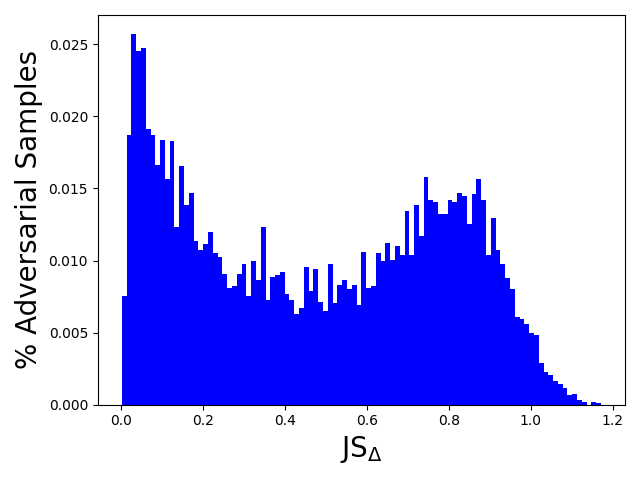} &
    \includegraphics[width=0.30\linewidth]{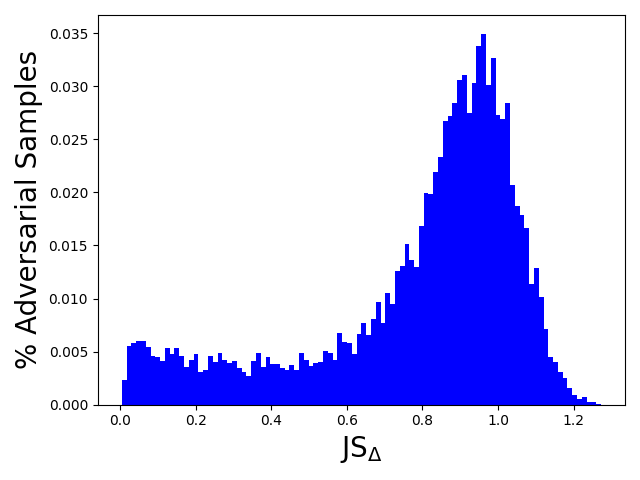} &
    \includegraphics[width=0.30\linewidth]{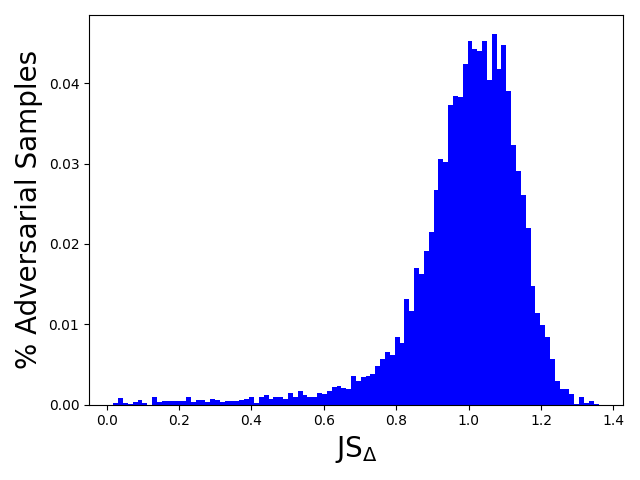} \\
  \end{tabular}
\caption{\small{Normalized histograms of $JS_\Delta$ for adversarial samples of a ResNet50 model trained via SAM ($\rho=0.3$) on CIFAR10. All samples are generated via a targeted PGD-100 attack. We observe a notable decline in adversarial bugs as compared to its SGD trained counterpart (Figure \ref{fig:cifar_hist}).}}
  \label{fig:cifar_sam_hist}
\end{figure*}

In Figures \ref{fig:cifar_sam_hist} and \ref{fig:svhn_sam_hist} and Tables \ref{tab:cifar_sam_comp} and \ref{tab:svhn_sam_comp}, we probe the effect of SAM on the makeup of adversarial samples for CIFAR10 and SVHN. We observe a notable rightward shift of the distribution of $JS_\Delta$ and a sharp decline in the percentage of highly non-transferable adversarial examples. Therefore, it appears that SAM offers protection against adversarial bugs. Similar to \cite{duality}, we also note a non-trivial robustness at small attack perturbations. 

To measure the effect of SAM on the robustness of the learned data features, we desire the minimum perturbation needed to induce data features, $\epsilon_f$, in models trained by SAM compared to standard and adversarial trained models. For this goal, we re-purpose fast minimum-norm (FMN) attack \cite{fmn} such that the gradient information comes from the ensemble $\tilde{f}$ and attack success is evaluated on the model of interest $f_0$. As shown in Table \ref{tab:fmn}, we observe that the minimum perturbation needed to induce data features for models trained with SAM is comparable to that of standard models, not of their robust counterparts. Therefore, contrary to the hypothesis of \cite{duality}, it appears that this regularization does \textit{not} encourage deep neural networks to utilize more robust features. Instead, it asymmetrically targets the weakness underlying adversarial bugs while leaving the robustness of features largely unchanged. This insight offers an explanation for the pathological differences between SAM and adversarial training noted by \cite{duality}, such as their effect on clean accuracy. It also suggests that the sharpness of a model's decision boundary plays a fundamental role in the existence of adversarial bugs.

\begin{table}[!h]
\caption{\small{Percentage of adversarial samples with $JS_\Delta<\beta$ for $\beta=[0.01, 0.05, 0.10]$ for a  ResNet50 model trained via SAM ($\rho=0.3$) on CIFAR10. All samples are generated via a targeted PGD-100 attack. We observe a notable decline in adversarial bugs as compared to its SGD trained counterpart (Table \ref{tab:cifar_comp}).}}
\label{tab:cifar_sam_comp}
\centering
\begin{tabular}{ccccc}
\toprule
\multirow{2}{*}{\begin{tabular}[c]{@{}c@{}}Attack \\ strength ($\epsilon$)\end{tabular}} & \multicolumn{1}{c}{\multirow{2}{*}{\begin{tabular}[c]{@{}c@{}}Robust \\ Accuracy\end{tabular}}} & \multicolumn{3}{c}{\% samples with $JS_\Delta<$}                                  \\ \cmidrule(l){3-5} 
                                                                                         & \multicolumn{1}{c}{}                                                                            & \multicolumn{1}{c}{$0.01$} & \multicolumn{1}{c}{$0.05$} & \multicolumn{1}{c}{$0.10$} \\
\midrule
    1 / 255 & 79.1\% & 0.5\%& 9.2\%& 27.8\%\\
    3 / 255 & 7.5\% & 0.5\% & 8.0\% & 16.2\%\\
    5 / 255 & 0.7\% & 0.1\%& 1.8\%& 3.8\%\\
    8 / 255 & 0.6\% & 0.0\% & 0.0\%& 0.2\%\\
    \bottomrule
    \end{tabular}
\end{table}

\begin{table}[!h]
\caption{\small{Average minimum magnitude of adversarial perturbation, $\|\epsilon_f\|_\mathit{p}$ where $\|\cdot\|_\mathit{p}$ denotes $L_p$ norm, needed to sufficiently manipulate data features  for models trained on CIFAR10 (left) and SVHN (right). We denote adversarially trained models as $M_\mathit{adv}$ and models trained using SAM $M_\rho$, where $\rho$ indicates the strength of SAM. Our results show that SAM does not increase the robustness of data features learned by our models.}}
\label{tab:fmn}
\adjustbox{max width=\columnwidth}{%
\begin{tabular}{@{}ccccc@{}}
\toprule
                  & \multicolumn{2}{c}{CIFAR10 ($\alpha=0.3$)} & \multicolumn{2}{c}{SVHN ($\alpha=0.1$)}      \\ \cmidrule(l){2-3}  \cmidrule(l){4-5} 
Model             & $\|\epsilon_f\|_2$       & $\|\epsilon_f\|_\infty$      & $\|\epsilon_f\|_2$ & $\|\epsilon_f\|_\infty$ \\ \midrule
$M_{\rho=0}$      & 0.38 $\pm$ 0.21          & 0.013 $\pm$ 0.006            & 0.65 $\pm$ 0.44    & 0.021 $\pm$ 0.013       \\
$M_{\rho=\alpha}$ & 0.38 $\pm$ 0.19          & 0.013 $\pm$ 0.006            & 0.59 $\pm$ 0.41    & 0.019 $\pm$ 0.011       \\
$M_{adv}$         & 1.61 $\pm$ 0.81          & 0.051 $\pm$ 0.024            & 1.03 $\pm$ 0.64    & 0.045 $\pm$ 0.024       \\ \bottomrule
\end{tabular}
}
\end{table}

\section{Re-examining Robust Datasets} \label{sec:gap}

Having confirmation that both non-robust features and adversarial bugs play a role in adversarial vulnerability, we might wonder whether the existence of the latter could explain the robustness gap observed between adversarially trained models and models trained on robust datasets. Namely, whether training on robust datasets does not offer protection against adversarial bugs. Indeed, we note that \cite{bugsrebut} used the robustness gap to hypothesize that robust features alone are insufficient for true model robustness. Adversarial bugs could offer an explanation for this behavior.

\begin{figure*}[!h]
\centering
\begin{tabular}{cccc}
    $D_\mathit{NR}$ &
    \includegraphics[height=0.5in]{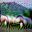} &
    \includegraphics[height=0.5in]{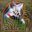} &
    \includegraphics[height=0.5in]{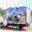}\\
    
    $D_\mathit{R}$ &
    \includegraphics[height=0.5in]{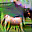} &
    \includegraphics[height=0.5in]{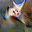} &
    \includegraphics[height=0.5in]{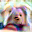}\\
    
    $D_\mathit{R'}$ &
    \includegraphics[height=0.5in]{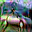} &
    \includegraphics[height=0.5in]{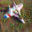} &
    \includegraphics[height=0.5in]{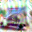}\\
    \\
    $(y_\mathit{src}, y_\mathit{target})$ & $(``car", ``horse")$ & $(``plane", ``cat")$ & $(``truck", ``dog")$\\
\end{tabular}
\caption{\small{(CIFAR10) Input samples from $D_{NR}$ (top row), $D_{R}$ (middle row), and a second-order robust dataset, $D_R'$, trained from $M_R$ (bottom row). We observe that images in $D_R'$ resemble their source images, as in $D_{NR}$, while only inputs in $D_R$ appear visually related to the target labels.}}
\label{fig:cifar_data_ex}
\end{figure*}

\begin{figure*}[!h]
  \centering
  \includegraphics[width=0.30\linewidth]{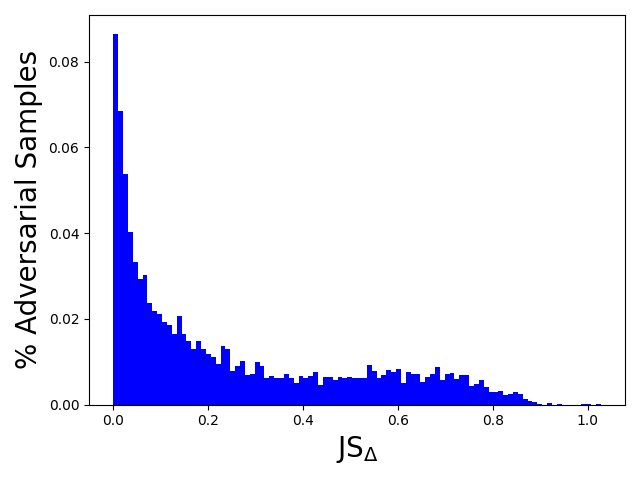} 
   \includegraphics[width=0.30\linewidth]{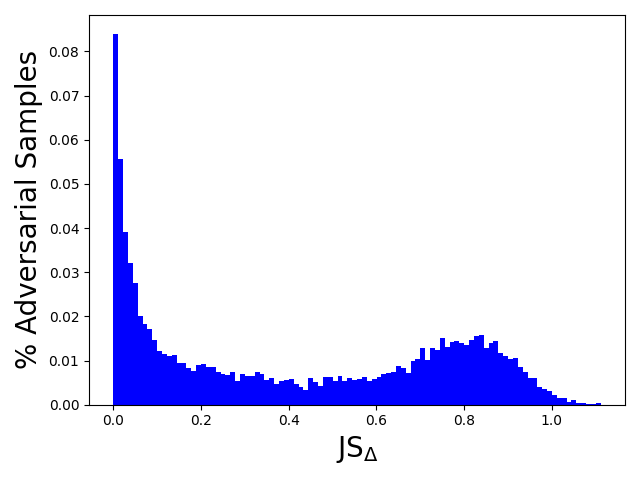} 
   \includegraphics[width=0.30\linewidth]{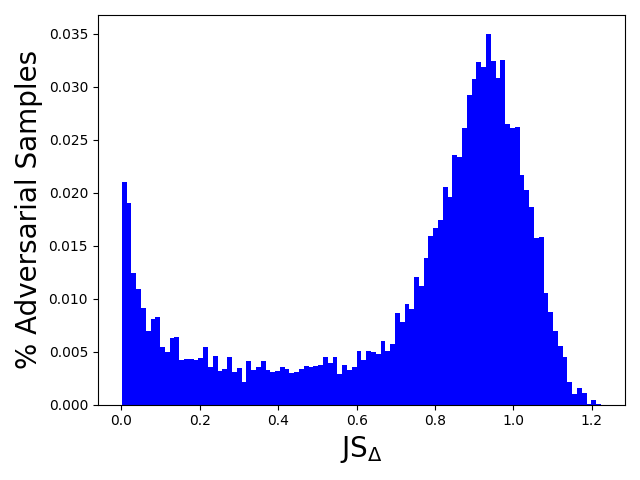} 
  \caption{\small{Normalized histograms of $JS_\Delta$ for adversarial samples of a ResNet50 model trained on a robust CIFAR10 dataset (from left to right, $\epsilon = 3/255, 5/255, 8/255$). All samples are generated via a targeted PGD-100 attack. We observe that the attacker is capable of manipulating data features at magnitudes comparable to standard-trained models, suggesting the re-emergence of non-robust features.}}
  \label{fig:cifar_r_hist}
\end{figure*}

In Figures \ref{fig:cifar_r_hist} and \ref{fig:svhn_r_hist}, we record $JS_\Delta$ histograms created by attacks on models, $M_R$, trained on robust CIFAR10 and SVHN datasets. Quantitative analyses of these histograms are provided in Tables \ref{tab:cifar_r_comp} and \ref{tab:svhn_r_comp}. Surprisingly, we observe perturbation magnitudes where the robustness gap exists, but the majority of successful adversarial samples for $M_R$ appear to utilize non-robust features instead of appearing as adversarial bugs (e.g. $\epsilon=8/255$). Thus, it appears that the robustness of features for $M_\mathit{R}$ has degraded in comparison to its adversarially trained counterpart. This suggests that the drop in robustness for these models is not due to a lack of protection against adversarial bugs but instead the result of a re-emergence of non-robust features. To confirm this possibility, we create a second-order robust dataset, $D_\mathit{R'}$, by applying the optimization procedure described in Equation \ref{eqn:dataset_gen} on the penultimate representation of $M_\mathit{R}$. In Figure \ref{fig:cifar_data_ex}, we visually inspect its images and compare them to those from $D_\mathit{NR}$ and $D_\mathit{R}$. We find that the images in $D_\mathit{R'}$ tend to resemble those in $D_\mathit{NR}$ markedly more than those in $D_\mathit{R}$. This suggests that the features learned by $M_\mathit{R}$ more closely align with those of standard trained models than adversarially trained -- that is, $M_\mathit{R}$ appears to be using incoherent, non-robust features.

\begin{table}[!h]
\caption{\small{Percentage of adversarial samples with $JS_\Delta<\beta$ for $\beta=[0.01, 0.05, 0.10]$ for a  ResNet50 model trained on a robust CIFAR10 dataset. All samples are generated via a targeted PGD-100 attack. We observe that the attacker is capable of manipulating data features at magnitudes comparable to standard-trained models, suggesting the re-emergence of non-robust features.}}
\label{tab:cifar_r_comp}
\centering
\begin{tabular}{ccccc}
    \toprule
\multirow{2}{*}{\begin{tabular}[c]{@{}c@{}}Attack \\ strength ($\epsilon$)\end{tabular}} & \multicolumn{1}{c}{\multirow{2}{*}{\begin{tabular}[c]{@{}c@{}}Robust \\ Accuracy\end{tabular}}} & \multicolumn{3}{c}{\% samples with $JS_\Delta<$}                                  \\ \cmidrule(l){3-5} 
                                                                                         & \multicolumn{1}{c}{}                                                                            & \multicolumn{1}{c}{$0.01$} & \multicolumn{1}{c}{$0.05$} & \multicolumn{1}{c}{$0.10$} \\
                                                                         
    \midrule
    3 / 255 & 42.2\% & 8.3\%& 27.8\%& 40.3\%\\
    5 / 255 & 9.3\% & 7.7\%& 22.6\%& 30.9\%\\
    8 / 255 & 2.0\% & 1.6\% & 6.4\%& 9.6\%\\
    \bottomrule
    \end{tabular}
\end{table}

Our finding is supported by \cite{invertcka, entangle}, which present evidence that the features of adversarially trained models are highly entangled with those of non-robust models and that robust features can be built via many useful non-robust features. However, to the best of our knowledge, its implications for training on robust datasets has not yet been appreciated. Because robust features contain non-robust useful features, it is often not possible to enforce their usage via standard training on robust datasets. This is due to the fact that features are not uniquely defined by the task (i.e. dataset) -- they are defined by \textit{how} the input is processed to solve the task. Therefore, we advocate that the robustness gap should not be viewed as evidence of an adversarial weakness that remains undefended when using robust features, as suggested by \cite{bugsrebut}. Instead, it is a direct consequence of naive attempts to distill them into a dataset. This insight also provides a natural explanation for why previous attempts to close this gap have been largely unsuccessful \cite{dataalone}.

\section{Conclusion}

In this paper, we emphasize that adversarial samples can be created via at least two distinct weaknesses -- model attention to non-robust features and a separate vulnerability that corresponds to sharpness in the loss landscape. We propose a metric to measure the usage of data features by adversarial samples based on their transferability to other model instances with different initializations. Using our metric, we provide the first analysis of the makeup of adversarial samples for non-robust and robust models with respect to their reliance on data features. Additionally, we demonstrate the usefulness of our perspective by clarifying two phenomena previously noted in literature - the effect of SAM on robustness and the inadequacy of robust datasets for robust learning. We believe that our contribution is a step forward towards understanding the pervasive yet nuanced vulnerability of neural networks to adversarial attacks.

{
    \small
    \bibliography{main}
}

\appendix
\onecolumn
\clearpage
\section{Replication of Experiments in Ilyas et al. \cite{bugs}} \label{app:replication}

In this section we provide the results of our reproduction of the experiments in \cite{bugs} on CIFAR10. In Table \ref{tab:robust_acc_cifar} we confirm that models trained on both $D_\mathit{NR}$ and $D_\mathit{R}$ result in non-trivial performance on the original test distribution. Our notation is as follows: $M$ is a model trained normally (i.e., SGD via cross-entropy loss), $M_\mathit{adv}$ is its adversarially trained counterpart, and $M_\mathit{NR}$ ($M_\mathit{R}$) denotes a model conventionally trained on $D_\mathit{NR}$ ($D_\mathit{R}$). Standard training only results in non-trivial accuracy, as compared to $M$, when using the robust dataset. However, similar to \cite{dataalone, bugsrebut}, we also note a robustness gap between $M_{R}$ and $M_{adv}$ (for example, 61.6\% $\rightarrow$ 2.0\% at $\epsilon=8/255$). 

In Table \ref{tab:det_rand}, we reproduce the experiment of \cite{bugs} using the non-robust datasets $D_\mathit{rand}$ and $D_\mathit{det}$. We confirm their finding that non-robust features are alone sufficient for learning, even in the presence of misaligned robust features, as in $D_\mathit{det}$.

\begin{table*}[!h]
\centering
\begin{tabular}{ccccc}
\toprule
\multirow{2}{*}{Model} & \multirow{2}{*}{Clean Accuracy} & \multicolumn{3}{c}{Robust Accuracy}\\
&  & $\epsilon=3/255$ & $\epsilon=8/255$ & $\epsilon=16/255$ \\
\midrule
$M$ & 94.1\%& 3.7\%& 0.7\%& 0.7\%\\
$M_\mathit{adv}$ & 87.8\% & 83.2\% & 61.6\% & 14.2\% \\
$M_\mathit{NR}$& 81.5\% & 3.4\% & 2.0\% & 2.0\%\\
$M_\mathit{R}$& 84.5\%&42.2\%& 2.0\%& 1.7\%\\
\midrule
\end{tabular}
\caption{\small{Replication of experiment in \cite{bugs} on CIFAR10, in which models conventionally trained on non-robust (robust) datasets display trivial (non-trivial) adversarial robustness. Robust accuracies are reported using a targeted $l_\infty$-constrained PGD-100 attack.}}
\label{tab:robust_acc_cifar}
\end{table*}

\begin{table}[!h]
\centering
\begin{tabular}{cc}
\toprule
Train Set & Clean Accuracy on $D$ \\
\midrule
$D_\mathit{rand}$ & 62.0\%\\
$D_\mathit{det}$ & 33.4\% \\
\midrule
\end{tabular}
\caption{\small{Replication of experiment in \cite{bugs} that shows nontrivial test accuracy on $D$ for models trained on the non-robust CIFAR10 datasets, $D_\mathit{rand}$ and $D_\mathit{det}$. This shows that models are capable of learning non-robust features even in the presence of misaligned robust features, as in $D_\mathit{det}$.}}
\label{tab:det_rand}
\end{table}

\section{Empirical Link between $JS_\Delta$ and the Generalization Property} \label{app:gen_exp}

To directly create adversarial bugs, \cite{discuss} proposes an ensemble-adjusted targeted attack loss:
\begin{align}
    \triangledown_{\boldsymbol{x}}L_{CE}(f_{\theta_0}(\boldsymbol{x}), y_\mathit{target}) \rightarrow \triangledown_{\boldsymbol{x}}L_{CE}(f_{\theta_0}(\boldsymbol{x}), y_\mathit{target})  + \triangledown_{\boldsymbol{x}}L_{CE}(\tilde{f}(\boldsymbol{x}), y_\mathit{src}).
    \label{eqn:ens_pgd}
\end{align}
Note that the second term in Equation \ref{eqn:ens_pgd} encourages small values of $JS_\Delta$ as it seeks to minimize $JS(\tilde{f}(\boldsymbol{x}_\mathit{src})|| \tilde{f}(\boldsymbol{x}))$. We use this ensemble-adjusted loss to create a second version of the non-robust CIFAR10 dataset, $D_\mathit{det}$, which we will denote as $D^\perp_\mathit{det}$. For the sake of completeness, we also construct an analogous version of $D_\mathit{rand}$ for CIFAR10, denoted here as $D^\perp_\mathit{rand}$. In Table \ref{tab:det_rand_perp} we record the resulting clean test accuracy when training on these datasets and observe that the test accuracy is notably lower than for $D_\mathit{rand}$ and $D_\mathit{det}$, as was observed in Table \ref{tab:det_rand}. Thus, we confirm the finding of \cite{discuss} that the transferability of adversarial samples to $\tilde{f}$ (and therefore $JS_\Delta$) is correlated with the Generalization property.

\begin{table}[!h]
\centering
\begin{tabular}{cc}
\toprule
Train Set & Test Accuracy on D \\
\midrule
$D^{\perp}_\mathit{rand}$ & 11.3\%\\
$D^{\perp}_\mathit{det}$ & 7.8\%\\
\midrule
\end{tabular}
\caption{\small{(CIFAR10) Clean test accuracy for models trained on ensemble-adjusted versions of $D_\mathit{det}$ and $D_\mathit{rand}$ that are constructed with Equation \ref{eqn:ens_pgd} (denoted here as $D^{\perp}_\mathit{det}$ and $D^{\perp}_\mathit{rand}$). We confirm the finding of \cite{discuss} that training on these datasets, which are comprised of highly nontransferable adversarial samples, does not result in generalization to $D$. This is in contrast to the results in Table \ref{tab:det_rand} of training on $D_\mathit{det}$ and $D_\mathit{rand}$.}}
\label{tab:det_rand_perp}
\end{table}

\section{Training Details} \label{app:training}

In this section, we provide training details for our models.

\paragraph{CIFAR10} For the ResNet50 architecture used for CIFAR10, we train using SGD for 200 epochs using momentum of 0.9 and an initial learning rate of 0.1. The learning rate is decayed by a factor of 0.1 every 50 epochs and we apply a weight decay factor of $0.0005$. We use a standard augmentation strategy that includes random horizontal flips, random cropping, small random rotations ($\leq2\degree$), and a random color jitter. We use these same settings for all the CIFAR10 models in our main results, including those that are adversarially trained, trained using SAM (with SGD remaining as the base optimizer), as well as trained on the robust CIFAR10 dataset.

\paragraph{SVHN} We use the ResNet18 architecture for SVHN and train using SGD for 20 epochs using momentum of 0.9 and an initial learning rate of 0.01. The learning rate is decayed by a factor of 0.1 at epoch 10 and we apply a weight decay factor of $0.0005$. For our augmentation strategy, we use random cropping as well as small random rotations ($\leq8\degree$). We use these same settings for all the SVHN models in Appendix \ref{app:svhn}, including those that are adversarially trained, trained using SAM (with SGD remaining as the base optimizer), as well as trained on the robust SVHN dataset.

\paragraph{Adversarial Training} When generating adversarially trained models for each dataset, we use an $\ell_\infty$-constrained PGD attack with a maximum perturbation of $\epsilon=8/255$. Our attack is performed over $7$ steps with a step size of $\epsilon/5$.

\section{SVHN Results} \label{app:svhn}

\begin{figure*}[!h]
\centering
\begin{tabular}{ccc}
    \small $\epsilon = 3/255$ &
    \small $\epsilon = 5/255$ &
    \small $\epsilon = 8/255$ \\
    \includegraphics[width=0.3\textwidth]{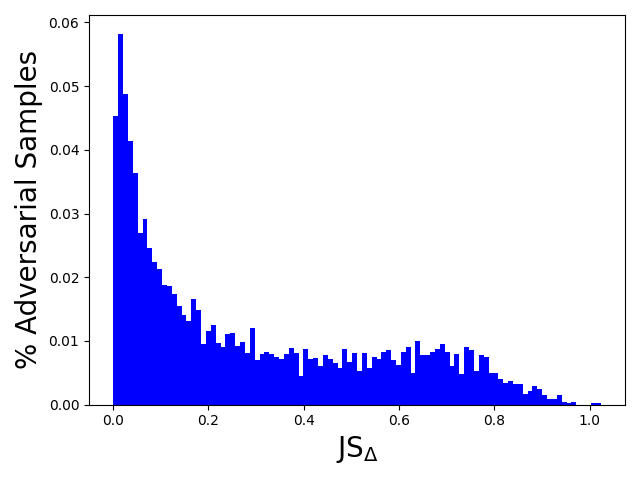} &
    \includegraphics[width=0.3\textwidth]{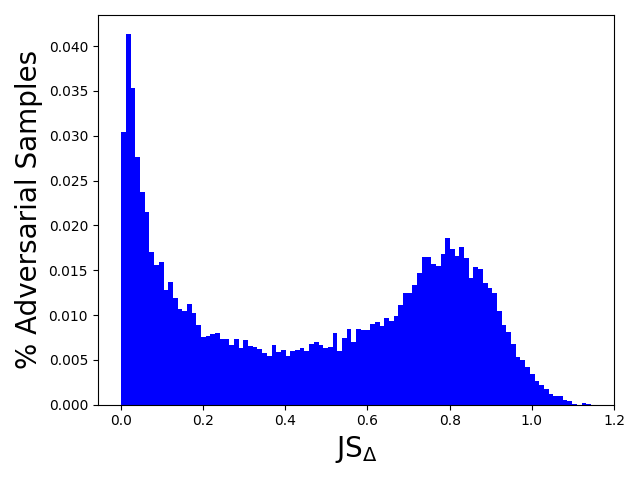} &
    \includegraphics[width=0.3\textwidth]{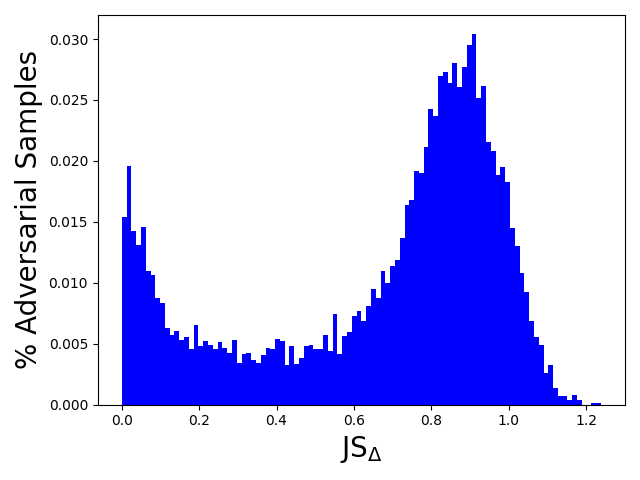} \\
\end{tabular}
\caption{\small{Normalized histograms of $JS_\Delta$ for successful adversarial samples of a non-robust ResNet18 model trained on SVHN. We observe that for low magnitudes of perturbation, a large percentage of adversarial samples can be identified as adversarial bugs. For larger magnitudes of perturbation, the attacker is able to manipulate non-robust features in the majority of adversarial samples.}}
\label{fig:svhn_hist}
\end{figure*}

\begin{table*}[!h]
\centering
\begin{tabular}{ccccc}
\toprule
\multirow{2}{*}{\begin{tabular}[c]{@{}c@{}}Attack \\ strength ($\epsilon$)\end{tabular}} & \multicolumn{1}{c}{\multirow{2}{*}{\begin{tabular}[c]{@{}c@{}}Robust \\ Accuracy\end{tabular}}} & \multicolumn{3}{c}{\% samples with $JS_\Delta<$}                                  \\ \cmidrule(l){3-5} 
                                                                                         & \multicolumn{1}{c}{}                                                                            & \multicolumn{1}{c}{$0.01$} & \multicolumn{1}{c}{$0.05$} & \multicolumn{1}{c}{$0.10$} \\
\midrule
3 / 255 & 47.7\% & 0.4\% & 22.3\%& 34.7\%\\
5 / 255 & 17.5\% & 0.2\% & 13.6\% & 21.5\%\\
8 / 255 & 4.0\% & 0.1\% & 0.6\%& 10.3\%\\
16 / 255 & 0.7\% & 0.0\% & 0.7\% & 1.2\% \\
\end{tabular}
\caption{\small{Percentage of adversarial samples with $JS_\Delta<\beta$ for $\beta=[0.01, 0.05, 0.10]$ for a non-robust ResNet18 model trained on SVHN. All samples generated via a targeted PGD-100 attack. We find that weak attacks tend to utilize adversarial bugs whereas a significant portion of adversarial samples use non-robust features beyond a certain threshold of attack perturbation.}}
\label{tab:svhn_comp}
\end{table*}

\begin{figure*}[!h]
\centering
\begin{tabular}{ccc}
    \small $\epsilon = 8/255$ &
    \small $\epsilon = 16/255$ &
    \small $\epsilon = 32/255$ \\
    \includegraphics[width=0.3\textwidth]{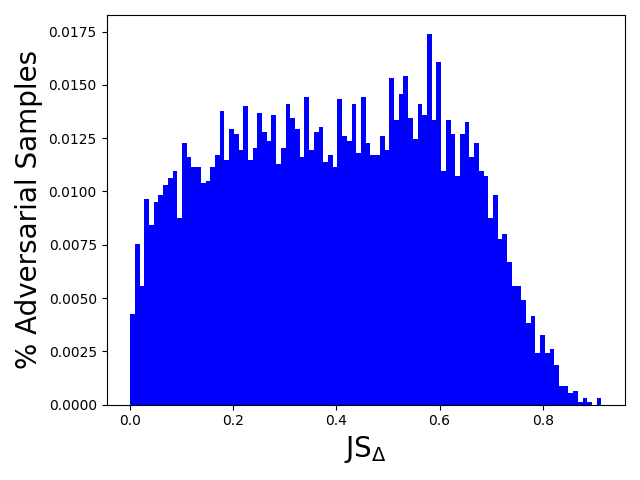} &
    \includegraphics[width=0.3\textwidth]{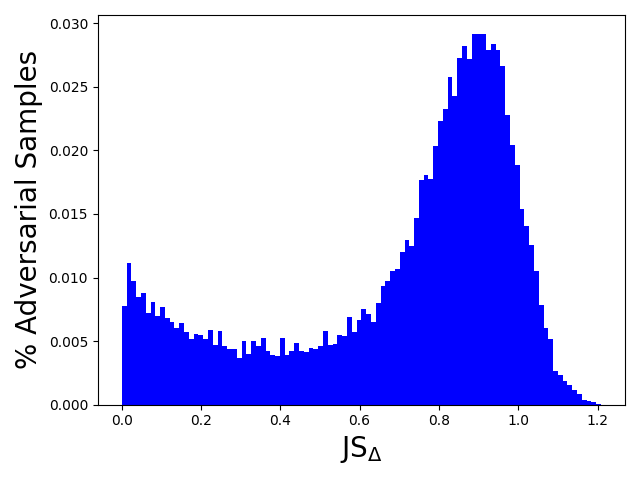} &
    \includegraphics[width=0.3\textwidth]{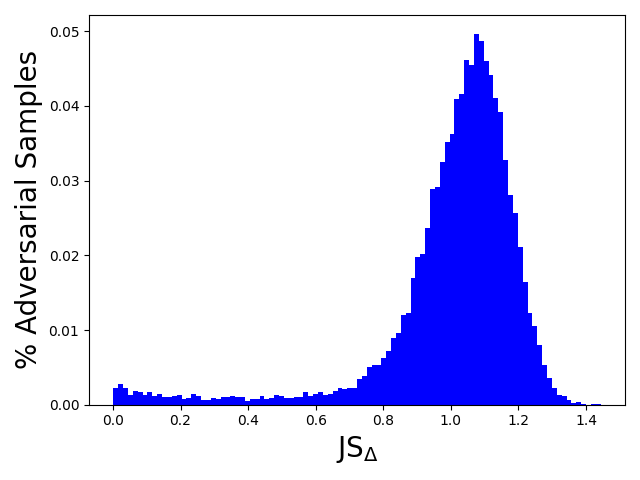}  \\
\end{tabular}
\caption{\small{Normalized histograms of $JS_\Delta$ for successful adversarial samples of a robust ResNet18 model that was adversarially trained on SVHN. Samples generated using a targeted PGD-100 attack. We observe that larger attack perturbations are needed to create adversarial samples that utilize non-robust features as well as a marked disappearance of adversarial bugs.}}
\label{fig:svhn_adv_hist}
\end{figure*}

\begin{table*}[!h]
\centering
\begin{tabular}{ccccc}
\toprule
\multirow{2}{*}{\begin{tabular}[c]{@{}c@{}}Attack \\ strength ($\epsilon$)\end{tabular}} & \multicolumn{1}{c}{\multirow{2}{*}{\begin{tabular}[c]{@{}c@{}}Robust \\ Accuracy\end{tabular}}} & \multicolumn{3}{c}{\% samples with $JS_\Delta<$}                                  \\ \cmidrule(l){3-5} 
                                                                                         & \multicolumn{1}{c}{}                                                                            & \multicolumn{1}{c}{$0.01$} & \multicolumn{1}{c}{$0.05$} & \multicolumn{1}{c}{$0.10$} \\
\midrule
5 / 255 & 82.1\% & 0.6\% & 5.2\% & 14.2\% \\
8 / 255 & 63.2\% & 0.4\% & 3.9\%& 9.4\%\\
16 / 255 & 15.9\% & 0.6\% & 3.8\%& 7.0\%\\
32 / 255 & 1.3\% & 0.1\% & 0.8\%& 1.3\%\\
\end{tabular}
\caption{\small{Percentage of adversarial samples with $JS_\Delta<\beta$ for $\beta=[0.01, 0.05, 0.10]$ for a robust ResNet18 model adversarially trained on SVHN. All samples generated via a targeted PGD-100 attack. We observe that larger attack perturbations are needed to create adversarial samples that utilize non-robust features as well as a marked disappearance of adversarial bugs.}}
\label{tab:svhn_adv_comp}
\end{table*}

\begin{figure*}[!h]
\centering
  \begin{tabular}{ccc}
  \small $\epsilon = 3/255$ &
    \small $\epsilon = 5/255$ &
    \small $\epsilon = 8/255$ \\
    \includegraphics[width=0.30\textwidth]{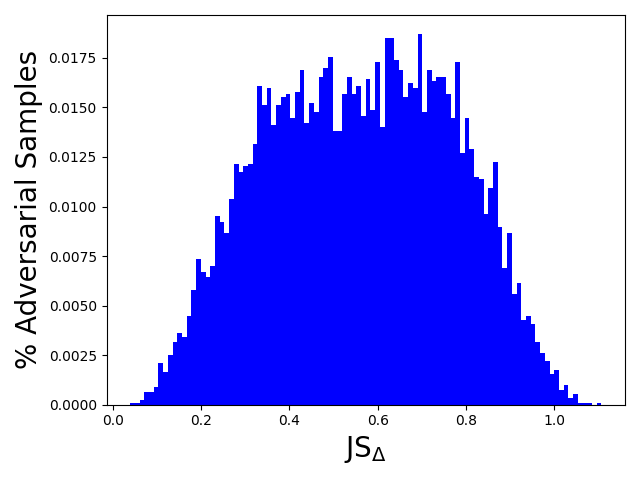} &
    \includegraphics[width=0.30\linewidth]{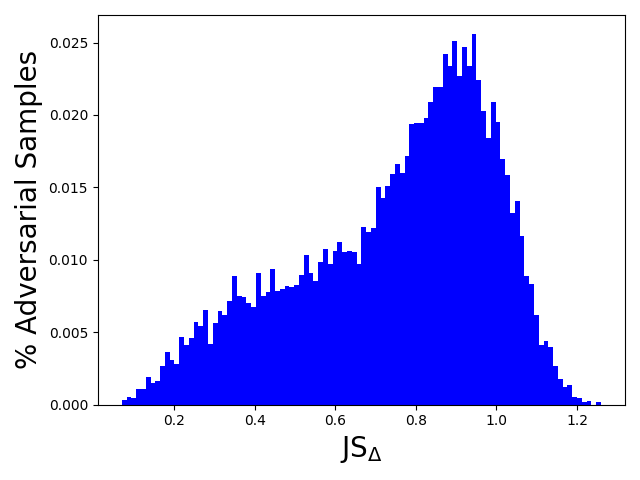} &
    \includegraphics[width=0.30\linewidth]{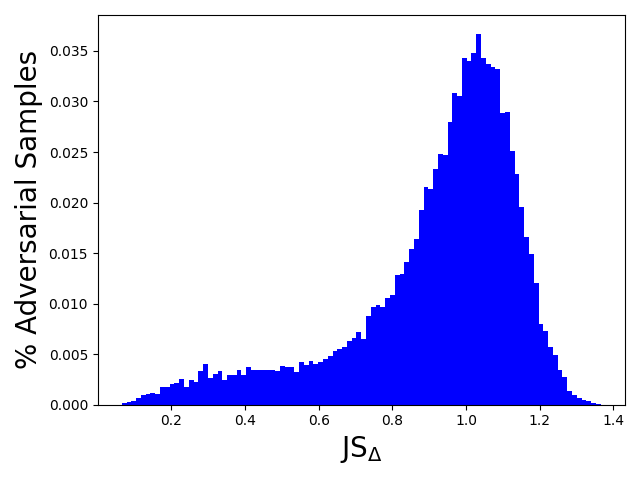} \\
  \end{tabular}
  \caption{\small{Normalized histograms of $JS_\Delta$ for successful adversarial samples of a ResNet18 model trained via SAM ($\rho=0.1$) on SVHN. Samples generated using a targeted PGD-100 attack. We observe a notable decline in adversarial bugs as compared to its SGD trained counterpart (Figure \ref{fig:svhn_hist}).}}
  \label{fig:svhn_sam_hist}
\end{figure*}

\begin{table*}[!h]
\centering
\begin{tabular}{ccccc}
    \toprule
     \multirow{2}{*}{\begin{tabular}[c]{@{}c@{}}Attack \\ strength ($\epsilon$)\end{tabular}} & \multicolumn{1}{c}{\multirow{2}{*}{\begin{tabular}[c]{@{}c@{}}Robust \\ Accuracy\end{tabular}}} & \multicolumn{3}{c}{\% samples with $JS_\Delta<$}                                  \\ \cmidrule(l){3-5} 
                                                                                         & \multicolumn{1}{c}{}                                                                            & \multicolumn{1}{c}{$0.01$} & \multicolumn{1}{c}{$0.05$} & \multicolumn{1}{c}{$0.10$} \\
    \midrule
    3 / 255 & 64.0\% & 0.0\% & 0.0\% & 0.2\%\\
    5 / 255 & 29.4\% & 0.0\%& 0.0\%& 0.1\%\\
    8 / 255 & 8.3\% & 0.0\% & 0.0\%& 0.0\%\\
    \end{tabular}
\caption{\small{Percentage of adversarial samples with $JS_\Delta<\beta$ for $\beta=[0.01, 0.05, 0.10]$ for a  ResNet18 model trained via SAM ($\rho=0.1$) on SVHN. Samples were generated using a targeted PGD-100 attack. We observe a notable decline in adversarial bugs as compared to its SGD trained counterpart (Table \ref{tab:svhn_comp}).}}
\label{tab:svhn_sam_comp}
\end{table*}

\begin{figure*}[!h]
  \centering
  \begin{tabular*}{\textwidth}{@{\extracolsep{\fill}}ccc}
  \small $\epsilon = 3/255$ &
    \small $\epsilon = 5/255$ &
    \small $\epsilon = 8/255$ \\
    \includegraphics[width=0.30\textwidth]{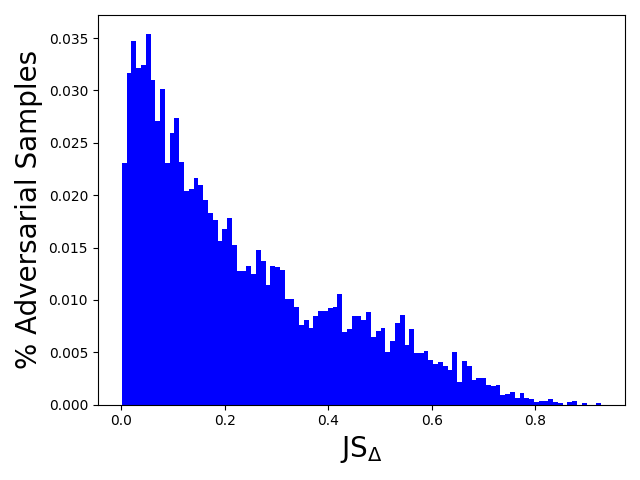} &
    \includegraphics[width=0.30\linewidth]{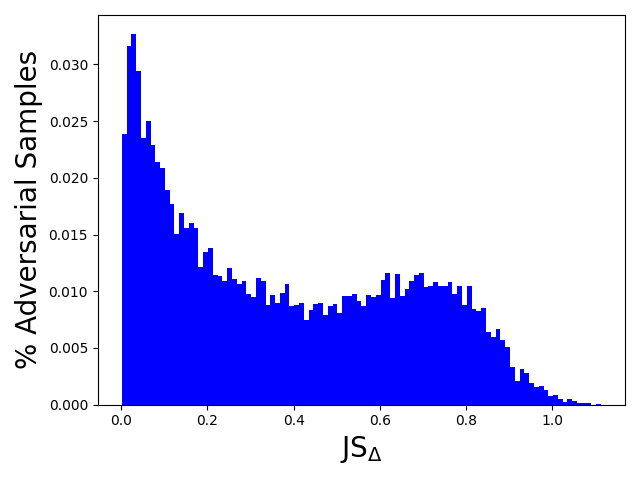} &
    \includegraphics[width=0.30\linewidth]{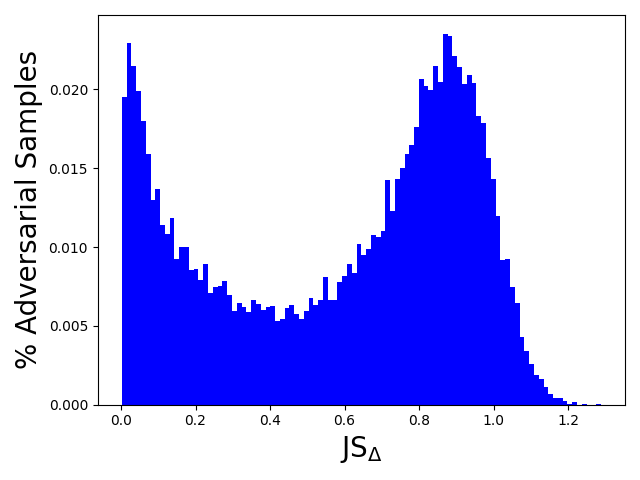} \\
  \end{tabular*}
  \caption{\small{Normalized histograms of $JS_\Delta$ for successful adversarial samples (generated via a targeted PGD-100 attack) of a ResNet18 model trained on a robust SVHN dataset. We observe that the attacker is capable of manipulating data features at magnitudes comparable to standard-trained models, suggesting the re-emergence of non-robust features.}}
  \label{fig:svhn_r_hist}
\end{figure*}

\begin{table*}[!h]
\centering
\begin{tabular}{ccccc}
    \toprule
     \multirow{2}{*}{\begin{tabular}[c]{@{}c@{}}Attack \\ strength ($\epsilon$)\end{tabular}} & \multicolumn{1}{c}{\multirow{2}{*}{\begin{tabular}[c]{@{}c@{}}Robust \\ Accuracy\end{tabular}}} & \multicolumn{3}{c}{\% samples with $JS_\Delta<$}                                  \\ \cmidrule(l){3-5} 
                                                                                         & \multicolumn{1}{c}{}                                                                            & \multicolumn{1}{c}{$0.01$} & \multicolumn{1}{c}{$0.05$} & \multicolumn{1}{c}{$0.10$} \\
    \midrule
    3 / 255 & 66.3\% & 2.2\%& 16.4\%& 31.9\%\\
    5 / 255 & 37.5\% & 1.5\%& 12.6\%& 22.9\%\\
    8 / 255 & 12.7\% & 1.0\% & 8.0\%& 14.0\%\\
    16 / 255 & 1.6\% & 0.1\% & 1.2\%& 2.2\%\\
\end{tabular}
\caption{\small{Percentage of adversarial samples with $JS_\Delta<\beta$ for $\beta=[0.01, 0.05, 0.10]$ for a  ResNet18 model trained on a robust SVHN dataset. All samples generated via a targeted PGD-100 attack. We observe that the attacker is capable of manipulating data features at magnitudes comparable to standard-trained models, suggesting the re-emergence of non-robust features.}}
\label{tab:svhn_r_comp}
\end{table*}

\begin{figure*}[!h]
\centering
\begin{tabular}{cccccccc}
    $D_{NR}$ &
    \includegraphics[height=0.5in]{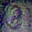} &
    \includegraphics[height=0.5in]{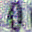} &
    \includegraphics[height=0.5in]{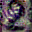}\\
    
    $D_R$ &
    \includegraphics[height=0.5in]{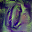} &
    \includegraphics[height=0.5in]{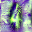} &
    \includegraphics[height=0.5in]{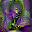}\\
    
    $D_R'$ &
    \includegraphics[height=0.5in]{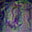} &
    \includegraphics[height=0.5in]{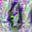} &
    \includegraphics[height=0.5in]{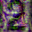}\\
    \\
    $(y_\mathit{src}, y_\mathit{target})$ & $(``3", ``0")$ & $(``0, ``4")$ & $(``2, ``9")$\\
\end{tabular}
\caption{\small{(SVHN) Input samples from $D_{NR}$ (top row), $D_{R}$ (middle row), and a second-order robust dataset, $D_R'$, trained from $M_R$ (bottom row). We observe that images in $D_R'$ resemble their source images, as in $D_{NR}$, while only inputs in $D_R$ appear visually related to the target labels.}}
\label{fig:svhn_data_ex}
\end{figure*}

\clearpage
\end{document}